\documentclass[10pt,twocolumn,letterpaper]{article}

\usepackage[pagenumbers]{iccv} 

\definecolor{iccvblue}{rgb}{0.21,0.49,0.74}
\usepackage[pagebackref,breaklinks,colorlinks,allcolors=iccvblue]{hyperref}

\def\paperID{7807} 
\def\confName{ICCV}
\def\confYear{2025}

\title{Improving Diffusion-Based Image Editing Faithfulness via Guidance and Scheduling}

\author{Hansam Cho\\
Korea University\\
{\tt\small chosam95@korea.ac.kr}
\and
Seoung Bum Kim\\
Korea University\\
{\tt\small sbkim1@korea.ac.kr}
}

\usepackage{multirow}
\usepackage{algorithm}
\usepackage{algorithmic}

\begin{document}

\twocolumn[{
\renewcommand\twocolumn[1][]{#1}
\maketitle
\vspace{-30pt}
\begin{center}
    \centering
    \captionsetup{type=figure}
    \includegraphics[width=0.95\linewidth]{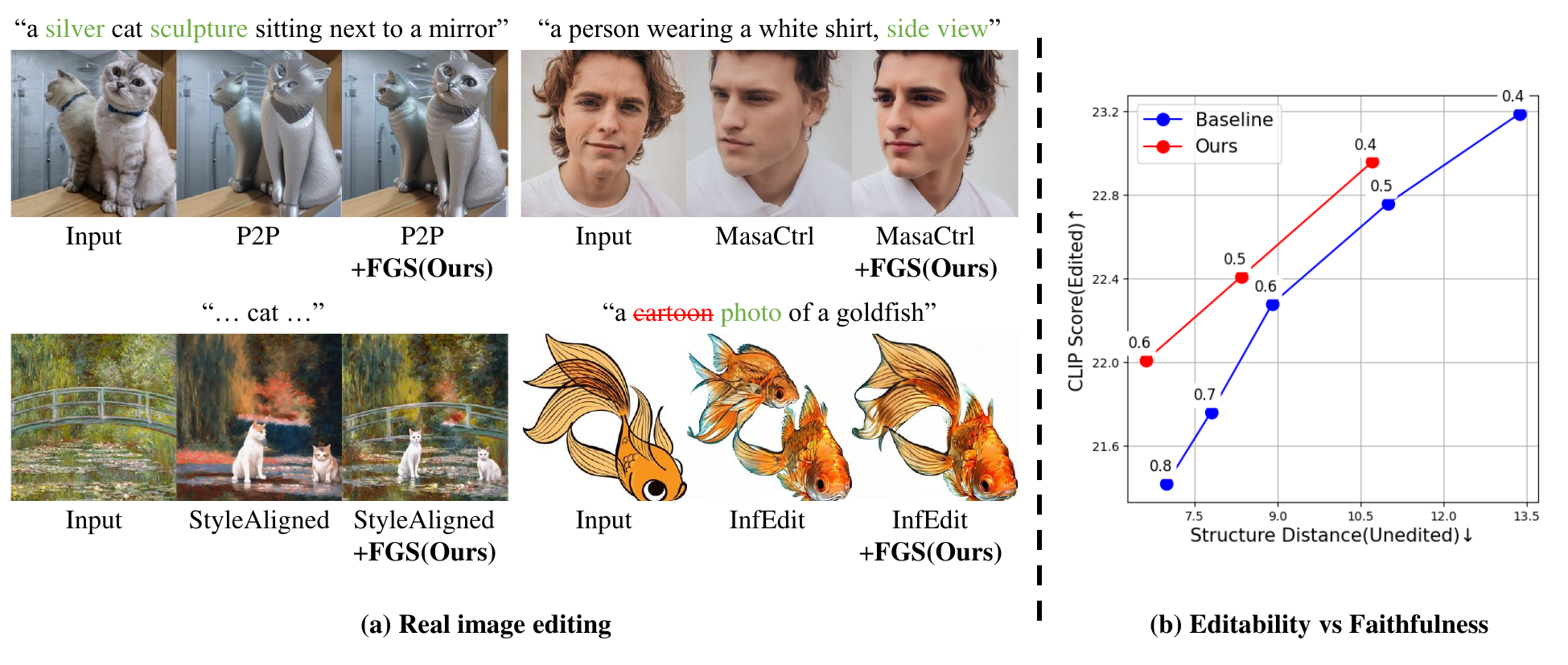}
    \vspace{-15pt}
    \caption{(a) Qualitative performance enhancement when combining FGS with real image editing techniques such as Prompt-to-Prompt (P2P)\cite{hertzprompt}, MasaCtrl\cite{cao2023masactrl}, StyleAligned~\cite{hertz2024style}, and InfEdit~\cite{xu2024inversion}. FGS more effectively preserves transferred information from input images, including content and style. The source text prompt is displayed above each image, with deleted words highlighted in red and additional words highlighted in green. (b) The editability-faithfulness trade-off relationship. Despite this trade-off, our FGS has achieved superior quantitative results.}
    \vspace{-5pt}
    \label{fig:1_teaser}    
\end{center}
}]

\maketitle
\begin{abstract}
Text-guided diffusion models have become essential for high-quality image synthesis, enabling dynamic image editing. In image editing, two crucial aspects are editability, which determines the extent of modification, and faithfulness, which reflects how well unaltered elements are preserved. However, achieving optimal results is challenging because of the inherent trade-off between editability and faithfulness. To address this, we propose Faithfulness Guidance and Scheduling (FGS), which enhances faithfulness with minimal impact on editability. FGS incorporates faithfulness guidance to strengthen the preservation of input image information and introduces a scheduling strategy to resolve misalignment between editability and faithfulness. Experimental results demonstrate that FGS achieves superior faithfulness while maintaining editability. Moreover, its compatibility with various editing methods enables precise, high-quality image edits across diverse tasks.
\end{abstract}    
\section{Introduction}
\label{sec:introduction}
\vspace{-5pt}
Text-guided diffusion models have become a standard for high-quality image synthesis, offering flexibility across various domains, particularly in creative content generation~\cite{rombach2022high, ramesh2022hierarchical, podellsdxl, chenpixart}. These models synthesize images through a multi-step denoising process, gradually transforming a noisy latent variable into an image. Recent research has leveraged this iterative process, demonstrating that manipulating intermediate latent enables diverse image editing tasks without additional training~\cite{hertzprompt, parmar2023zero, cao2023masactrl, tumanyan2023plug, alaluf2024cross, hertz2024style}.

For real image editing with these techniques, an initial inversion step is required, where the input image is transformed into a latent variable~\cite{songdenoising}. This latent is then processed through two parallel pathways: a reconstruction path and an editing path~\cite{hertzprompt, cao2023masactrl, tumanyan2023plug, xu2024inversion}. The reconstruction path reconstructs the input image based on a given source text prompt, while the editing path selectively modifies the image according to a new target prompt. By incorporating contextual information from the reconstruction path, the editing path can retain key details, enabling precise edits without compromising essential elements of the original image.

The information transferred from the reconstruction path is typically controlled by hyperparameters that select the specific denoising steps or particular layers within the diffusion model~\cite{hertzprompt,cao2023masactrl,tumanyan2023plug, xu2024inversion}. Adjusting these hyperparameters controls a trade-off between two critical aspects: \textit{editability}, which determines the extent of the image modification, and \textit{faithfulness}, which ensures the preservation of unaltered elements. As illustrated in Fig.\ref{fig:1_teaser}(b), the x-axis represents structure distance (SD)\cite{tumanyan2022splicing} between the input and edited images, measuring faithfulness, while the y-axis indicates the CLIP score~\cite{hessel2021clipscore} between the edited image and target text, assessing editability. The hyperparameter values, shown above each point in Fig.\ref{fig:1_teaser}(b), control editability and faithfulness. However, increasing faithfulness (lower SD) typically reduces editability (lower CLIP score), and the opposite is also true. This inherent trade-off makes achieving an optimal edited image challenging, as improving one aspect often requires sacrificing the other.

To address this challenge, we propose Faithfulness Guidance and Scheduling (FGS), a novel approach designed to enhance faithfulness in real image editing. Our method introduces additional guidance to improve faithfulness, combined with a scheduling strategy that minimizes misalignment between faithfulness and editability. Together, FGS ensures high-quality image editing results. As shown in Fig.~\ref{fig:1_teaser}(b), even with the inherent trade-off, FGS achieves superior performance compared to the baseline, demonstrating lower SD and higher CLIP scores.

Our experimental results highlight FGS's ability to preserve the faithfulness components of the input image while seamlessly integrating with various editing methods. As illustrated in Fig.~\ref{fig:1_teaser}(a), baseline results fail to retain essential information from the input image, such as content and style. In contrast, FGS precisely preserves this transferred information while maintaining editing capabilities. Our main contributions are as follows:
\begin{itemize}
    \item We propose faithfulness guidance to improve faithfulness in real image editing.
    \item To mitigate the misalignment between faithfulness and editability, we introduce a scheduling method.
    \item We integrate these techniques into the Faithfulness Guidance and Scheduling (FGS), demonstrating that FGS enhances faithfulness while preserving editability across various editing methods.
\end{itemize}
\section{Related work}
\label{sec:related work}

\subsection{Image editing with diffusion models}
\vspace{-5pt}
Recent advancements in image editing have introduced several methods focused on maintaining essential attributes of images during the editing process. Some approaches preserve the spatial layout of the image while editing. Prompt-to-Prompt (P2P)\cite{hertzprompt} ensures that the overall layout of the image is preserved by transferring the attention map from the reconstruction path to the editing path. Similarly, Plug-and-Play\cite{tumanyan2023plug} focuses on retaining spatial details by transferring both the query and key of attention mechanisms, along with relevant features. Pix2pix-zero~\cite{parmar2023zero} uses cross-attention guidance to maintain the overall structure of the image during image-to-image translation.

In contrast to methods that emphasize layout preservation, MasaCtrl~\cite{cao2023masactrl} introduces a mutual self-attention mechanism for non-rigid editing. Given a text prompt, it generates the overall structure of the image, while the reconstruction path transfers the content of the input image to ensure consistency in the edited image. StyleAligned~\cite{hertz2024style} proposes a shared attention layer to transfer the style of the input image. InfEdit~\cite{xu2024inversion} introduces an inversion-free editing method for fast editing using consistency models~\cite{song2023consistency, luo2023latent} and additionally incorporates unified attention control to support both rigid and non-rigid editing. We demonstrate that our FGS can be seamlessly integrated with these prior editing methods to enhance the faithfulness of the editing results.

\subsection{Enhanced Techniques for Real Image Editing}
\vspace{-5pt}
To edit a real image, the input image first needs to be encoded into a latent variable using DDIM inversion~\cite{songdenoising}, then reconstructed to its original image via DDIM sampling~\cite{songdenoising} to transfer input image information effectively. However, applying these methods directly to real image editing can yield suboptimal results~\cite{wallace2023edict, mokady2023null}. To improve outcomes, some studies aim to correct errors that arise during the inversion process. For example, EDICT~\cite{wallace2023edict}, inspired by the affine coupling layer~\cite{dinh2014nice, dinh2016density}, introduces an auxiliary diffusion state to enable precise inversion. BDIA~\cite{zhang2025exact} employs a bidirectional structure in DDIM sampling to correct inversion errors, while AIDI~\cite{pan2023effective} utilizes fixed-point iteration to reduce errors during inversion.

Other approaches focus on reducing errors in the reconstruction process. NTI~\cite{mokady2023null}, for instance, optimizes the null text embedding to prevent collapse in the reconstruction path, while NMG~\cite{chonoise} applies guidance techniques to prevent the reconstruction path from diverging from the inversion path. Direct Inversion~\cite{ju2024direct} calculates the distance between the inversion and reconstruction paths, adjusting the reconstruction path based on the calculated error. Despite these improvements in inversion and reconstruction processes, these methods still face a trade-off between editability and faithfulness. We demonstrate that our method seamlessly integrates with these enhancement techniques, further improving editing quality.

\begin{figure*}\centering
    \includegraphics[width=\linewidth]{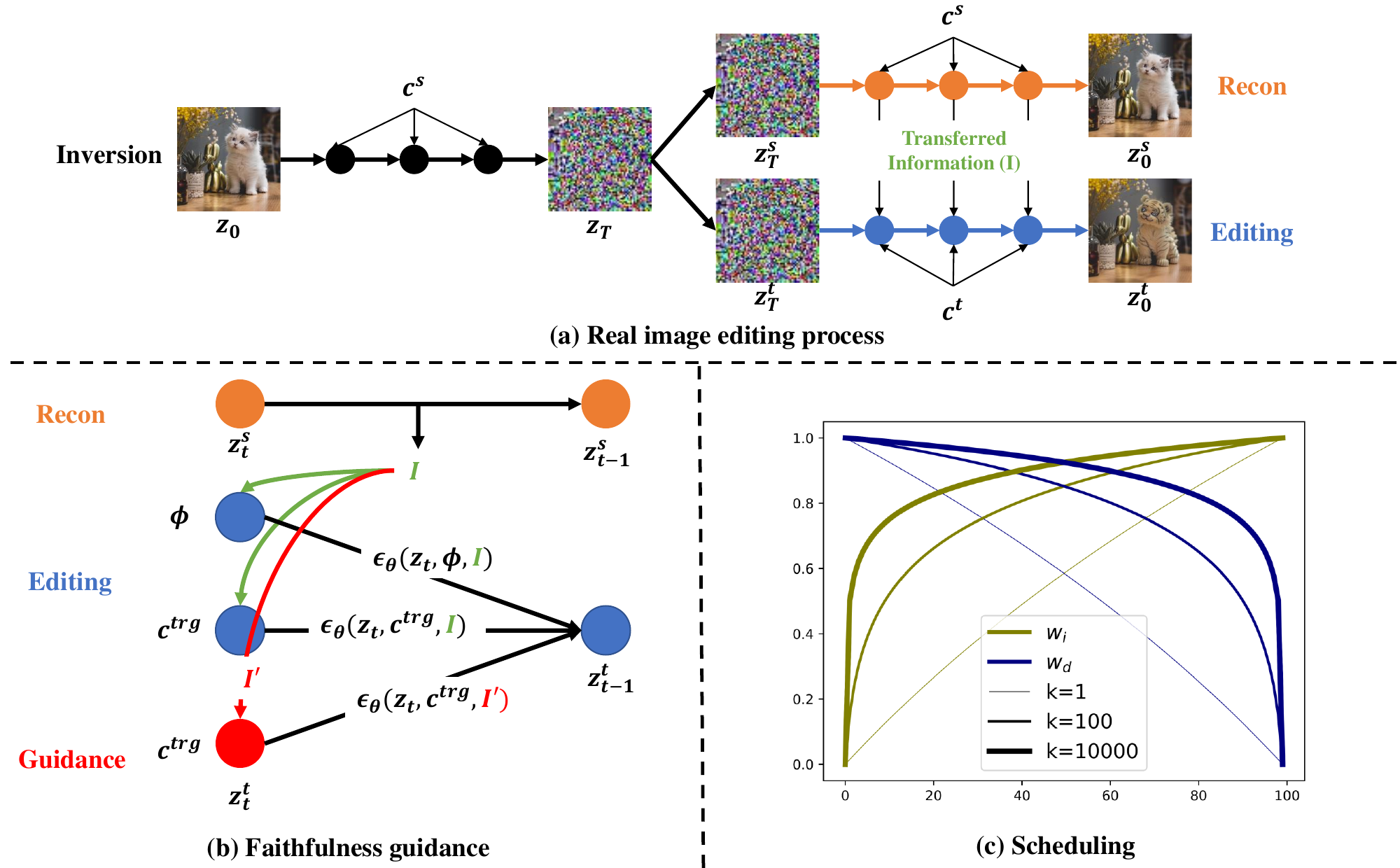}
    \vspace{-10pt}
    \caption{(a) Real image editing process with diffusion models. The editing process begins by inverting the input image $z_0$ into latent space, followed by two parallel paths: reconstruction and editing. Transferred information $I$ from the reconstruction path is incorporated into the editing path to enhance faithfulness. (b) Faithfulness guidance is applied by perturbing the transferred information $I$ to produce $I'$, which is then injected into the editing path to improve faithfulness during editing. (c) Visualization of the scheduling strategy for the two guidance scales, $w_d$ and $w_i$, for different values of $k$. As $k$ varies, the scheduling curve adjusts, balancing the emphasis between editability and faithfulness across timesteps.}
    \vspace{-10pt}
    \label{fig:2_method}
\end{figure*}

\subsection{Variation of classifier-free guidance}
Classifier guidance~\cite{dhariwal2021diffusion} was the first to introduce the guidance technique in diffusion models, using an external classifier to steer the generation process toward a desired condition. Classifier-free guidance (CFG)\cite{ho2021classifier} advanced this technique by eliminating the need for an additional classifier, using both conditional and unconditional models within the guidance process. However, CFG requires training both a conditional and an unconditional version of the model, which can be computationally demanding. To address this limitation, Autoguidance~\cite{karras2024guiding} proposed guiding the diffusion process using a smaller or less trained version of the model. Similarly, ICG~\cite{sadat2024no} eliminates the need for an unconditional model by employing conditions that are independent of the input image.

Moreover, CFG applies only to conditional models. To broaden its applicability, recent methods have focused on perturbing internal features of diffusion models to provide guidance, allowing these techniques to be applied to unconditional models as well. The main challenge in this approach is determining the optimal way to perturb the internal features of diffusion models. For example, SAG~\cite{hong2023improving} blurs regions containing salient information identified through self-attention maps, while PAG~\cite{ahn2024self} replaces the self-attention maps with an identity matrix. SEG~\cite{hong2024smoothed} shows that blurring the self-attention weights flattens the energy landscape, achieving a similar effect to CFG. Our approach is similar to PAG and SEG. However, rather than perturbing the model’s internal features, we focus on perturbing the transferred information from the reconstruction path to improve the faithfulness of image editing.
\section{Method}
\label{sec:method}

\subsection{Background}
Text-guided diffusion models~\citep{rombach2022high} are designed to map a random Gaussian noise vector $z_T$ to an image $z_0$ that aligns with a specified text condition $c$, typically represented by embeddings from text encoders such as CLIP~\citep{radford2021learning}. This mapping is achieved through a sequential denoising process, commonly referred to as the reverse process, driven by a noise prediction network $\epsilon_\theta$, which is optimized using the following loss function:
\vspace{-5pt}
\begin{equation} \label{eq:loss_simple}
    L_{simple} = E_{z_0,\epsilon \sim N(0,I),t \sim U(1,T)} \| {\epsilon - \epsilon_\theta(z_t, t, c)} \|_2 ^2.
\end{equation}

In real image editing with diffusion models, we begin by encoding the input image from $z_0$ to $z_T$ using DDIM Inversion~\cite{songdenoising}, defined by the equation:
\vspace{-5pt}
\begin{equation} \label{eq:ddim_inversion}
    z_{t+1}=\sqrt{\tfrac{\alpha_{t+1}}{\alpha_{t}}}z_t+\sqrt{\alpha_{t+1}} \left(A_{t+1} -A_t \right)\epsilon_{\theta}(z_t, c^{s})
\end{equation}
where $\{\alpha_t\}_{t=0}^T$ is a predefined noise schedule, $A_t = \sqrt{1/(\alpha{t} - 1)}$, and $c^{s}$ is a source text prompt that aligns with the input image. After encoding the latent $z_T$, we initiate two parallel paths: a reconstruction path and an editing path. The reconstruction path generates a reconstructed image $z_0^{s}$ from $z_T^{s}$, while the editing path produces an edited image $z_0^{t}$ from $z_T^{t}$. Both $z_T^{s}$ and $z_T^{t}$ are initialized from the same value $z_T$. The overall real image editing process is illustrated in Fig.~\ref{fig:2_method} (a).

In the reconstruction path, CFG~\cite{ho2021classifier} is applied, modifying the diffusion model output as follows: 
\vspace{-5pt}
\begin{equation} \label{eq:cfg_recon}
    \tilde{\epsilon}_\theta(z_t, c^{s}) = \epsilon_\theta\left(z_t, c^{s} \right) \!+\! w_{cfg}(\epsilon_\theta\left(z_t, c^{s} \right) \!-\! \epsilon_\theta\left(z_t, \phi \right) ).
\end{equation}
Using this modified output, denoising is performed through DDIM sampling~\cite{songdenoising} as follows:
\vspace{-5pt}
\begin{equation} \label{eq:ddim_reverse_recon}
    z_{t-1}^s=\sqrt{\tfrac{\alpha_{t-1}}{\alpha_{t}}}z_t + \sqrt{\alpha_{t-1}} \left(A_{t-1} - A_t \right)\tilde{\epsilon}_{\theta}(z_t, c^{s}).
\end{equation}
To transfer information from the reconstruction path to the editing path, each denoising operation in the reconstruction path generates input image information, which is then incorporated into the editing path. This allows the editing path to retain key details of the input image. The transferred information is integrated into the editing path as follows:
\vspace{-5pt}
\begin{align} \label{eq:cfg_edit}
    \tilde{\epsilon}_\theta(z_t, c^{t}, I)&=\epsilon_\theta\left(z_t, c^{t}, I \right) \nonumber \\
    &+ w_{cfg}\left(\epsilon_\theta\left(z_t, c^{t}, I \right) - \epsilon_\theta\left(z_t^t, \phi, I \right)\right) 
\end{align}
\vspace{-20pt}
\begin{align} \label{eq:ddim_reverse_edit}
    z_{t-1}^t=\sqrt{\tfrac{\alpha_{t-1}}{\alpha_{t}}}z_t^t + \sqrt{\alpha_{t-1}} \left(A_{t-1} - A_t \right)\tilde{\epsilon}_{\theta}(z_t^t,c^{t},I)
\end{align}
where $I$ represents the transferred information from the reconstruction path to the editing path, which varies depending on the editing technique. For example, in P2P~\cite{hertzprompt}, the transferred information is an attention map, while in MasaCtrl~\cite{cao2023masactrl}, it involves the key and value components of the attention mechanism.

\subsection{Faithfulness guidance}
To generate high-quality edited results within the constraints of the faithfulness-editability trade-off, we focus on improving faithfulness while maintaining a fixed level of editability. To this end, we introduce an additional guidance path incorporating perturbed information from the reconstruction path to enhance faithfulness in the editing process.

In diffusion models, the score function can be approximated as $\nabla_{z_t} \log p(z_t) \approx -\epsilon_{\theta}(z_t) / \sqrt{1 - \alpha_t}$. This score function can be incorporated into Eq.~\ref{eq:cfg_recon}, modifying the diffusion model’s conditional probability as follows:
\vspace{-5pt}
\begin{equation}
    \tilde{p_{\theta}}(z_t|c) = p_{\theta}(z_t|c)^{(1+w_{cfg})}p_{\theta}(z_t|\phi)^{-w_{cfg}}
\end{equation}
In CFG, $w_{cfg}$ is typically set to a value greater than one, resulting in a sharper distribution $\tilde{p_{\theta}}(z_t|c)$ compared to the original distribution $p_{\theta}(z_t|c)$.

Similarly, our goal is to enhance the information transferred from the reconstruction path to improve the faithfulness of the editing result. We treat the transferred information as a conditional input for the editing path, allowing us to apply a CFG-like guidance method to amplify faithfulness. However, since an explicit unconditional output is not available in this context, we take inspiration from previous research~\cite{ahn2024self, hong2024smoothed}, which shows that perturbing attention weights can effectively approximate an unconditional state.

We propose faithfulness guidance (FG), where we apply perturbations to the transferred information to approximate an unconditional output. Since this transferred information primarily consists of attention weights, perturbing these weights enables us to approximate an unconditional output. Similar to CFG, this approach allows us to amplify the transferred information as follows:
\vspace{-5pt}
\begin{align} \label{eq:perturb}
    I' := PERTURB(I)
\end{align}
\vspace{-20pt}
\begin{align}
    \tilde{\epsilon}_\theta(z_t^t, c^{t},I)&=\epsilon_\theta\left(z_t^t, c^t, I \right) \nonumber \\
    &+ w_{fg}\left(\epsilon_\theta\left(z_t^t, c^{t}, I \right) - \epsilon_\theta\left(z_t^t, c^{t}, I' \right)\right)
\end{align}
We use Gaussian blur as the perturbation method. By integrating this approach with the original editing path, we update Eq.~\ref{eq:cfg_edit} as follows:
\vspace{-5pt}
\begin{align} \label{eq:cfg_edit_fg}
    \tilde{\epsilon}_\theta(z_t^t,c^{t}, I) &= \epsilon_\theta(z_t^t,c^{t},I) \nonumber \\
    &+w_{cfg} (\epsilon_\theta(z_t^t,c^{t},I)-\epsilon_\theta(z_t^t,\phi,I)) \nonumber\\
    &+w_{fg} (\epsilon_\theta(z_t^t,c^{t},I)-\epsilon_\theta(z_t^t,c^{t},I'))
\end{align}
This combined formulation enables the editing path to incorporate both CFG and FG, enhancing the integration of transferred information and thereby improving faithfulness. The one-step denoising process of FG is illustrated in Fig.\ref{fig:2_method}(b).

While our approach shares some similarities with PAG~\cite{ahn2024self} and SEG~\cite{hong2024smoothed}, it diverges in a key aspect: instead of perturbing the internal features used for image synthesis, our approach perturbs transferred information from the reconstruction path for image editing. SEG demonstrated that Gaussian blur smooths the energy function landscape, approximating an unconditional output and improving synthesized image quality, similar to CFG. However, it does not selectively target specific conditions. In contrast, we empirically show that this technique can be extended to enhance specific conditions. Our method selectively perturbs transferred information, using the perturbed output as an unconditional reference, which enhances input image information and improves faithfulness.

\begin{figure}\centering
    \includegraphics[width=\linewidth]{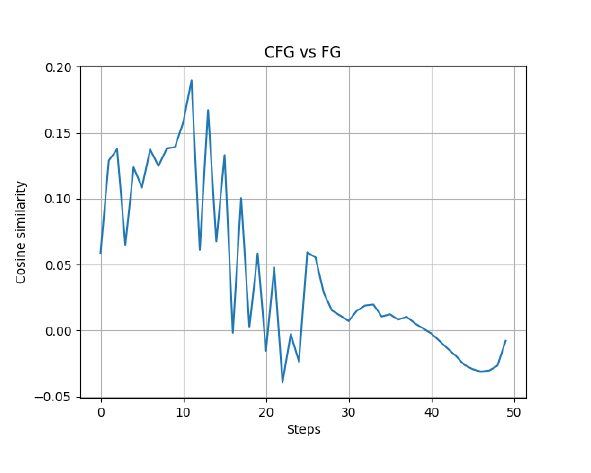}
    \vspace{-30pt}
    \caption{Cosine similarity between CFG and FG directions over timesteps. We edited 100 images and computed the average cosine similarity at each timestep.}
    \vspace{-10pt}
    \label{fig:3_misalignment}
\end{figure}

\subsection{Scheduling}
However, introducing FG alone is insufficient. We found that the original CFG direction, $\epsilon_\theta(z_t, c^{t}, I) - \epsilon_\theta(z_t, \phi, I)$, which represents editability, and the additional FG direction, $\epsilon_\theta(z_t, c^{t}, I) - \epsilon_\theta(z_t, c^{t}, I')$, which represents faithfulness, exhibit misalignment. Fig.~\ref{fig:3_misalignment} shows the cosine similarity over timesteps, revealing that the similarity remains consistently low as denoising progresses. This suggests that when both directions are combined, they struggle to perform their intended functions effectively.

To address this issue, we introduce a scheduling method to resolve the misalignment. Since diffusion models generate images in a coarse-to-fine manner~\cite{songdenoising, choi2021ilvr, choi2022perception}, it is essential to preserve the overall content of the input image in early timesteps (i.e., noisier latent) and then refine the image with finer details in later timesteps (i.e., denoised latent). Considering this, we adjust the emphasis on the transferred information on the timestep. For instance, if the transferred information relates to the overall layout of the image, as in P2P~\cite{hertzprompt}, we emphasize FG in the early timesteps. Conversely, if the information is relevant to fine details (e.g., texture or color), as in StyleAligned~\cite{hertz2024style}, we emphasize FG in the later timesteps.

We use a logarithmic function to ensure that one guidance scale decreases monotonically while the other increases, as defined below:
\vspace{-5pt}
\begin{gather}
    w_{t}^{d} = w^{d}\frac{log(1+k(T-t))}{log(1+k)}, w_{t}^{i} = w^{i}\frac{log(1+kt)}{log(1+k)}
\end{gather}
where $w_d$ decreases monotonically over the timesteps, emphasizing the early timesteps, while $w_i$ increases monotonically, emphasizing the later timesteps. As shown in Fig.~\ref{fig:2_method} (c), when the logarithmic function changes smoothly (i.e., with a smaller $k$ value), the two guidance scales do not conflict. However, if the function changes sharply (i.e., with a larger $k$ value), conflict may arise as both guidance scales remain higher across all timesteps. The parameter $k$ controls the rate of change, providing flexibility in balancing FG and CFG throughout the denoising process.

Finally, as the scheduling is introduced, we can update Eq.~\ref{eq:cfg_edit_fg} as follows:
\vspace{-5pt}
\begin{align} \label{eq:cfg_edit_fg_sch}
    \tilde{\epsilon}_\theta(z_t^t,c^{t}, I) &= \epsilon_\theta(z_t^t,c^{t},I) \nonumber \\
    &+w_{cfg}^t (\epsilon_\theta(z_t^t,c^{t},I)-\epsilon_\theta(z_t^t,\phi,I)) \nonumber\\
    &+w_{fg}^t (\epsilon_\theta(z_t^t,c^{t},I)-\epsilon_\theta(z_t^t,c^{t},I')).
\end{align}
As mentioned earlier, $w_{cfg}$ and $w_{fg}$ are set based on the type of transferred information. For instance, when transferring the layout of the image, as in P2P, $w_{fg}$ is set to $w_d$ and $w_{cfg}$ to $w_i$. Conversely, when transferring the style of the image, as in StyleAligned, the assignments are reversed.

Our full method, FGS, is shown in Alg.~\ref{alg:fgs}. First, we obtain the transferred information from the reconstruction path. We then perturb the transferred information $I$ to $I'$, incorporating it into the editing path. The editing path integrates the original transferred information $I$, the perturbed information $I'$, and the scheduled guidance scales $w_t^{cfg}$ and $w_t^{fg}$. By applying the modified diffusion model output to the sampling method, we obtain the one-step denoised latent. It is worth noting that the FGS process can be implemented in parallel with the original editing method, so our approach does not require additional time-consuming steps.

\begin{algorithm}[!t]    
    \caption{FGS}
        \textbf{Input:} source prompt $c^{s}$, target prompt $c^{t}$, inital source and target latent $z^{s}_T$ and $z^{t}_T$. \\
        \textbf{Output:} reconstructed image $z^{s}_0$, edited image $z^{t}_0$.
        
        \begin{algorithmic}[1]
        \FOR{$t = T, T-1, ..., 1$}
            \STATE $\tilde{\epsilon}_\theta(z_t^s, c^{s}), I \leftarrow RE(z_t^s, c^s, w_{cfg})$ \hfill Eq.~\ref{eq:cfg_recon}
            \STATE $I':=PERTURB(I)$ \hfill Eq.~\ref{eq:perturb}
            \STATE $\tilde{\epsilon}_\theta(z_t^t,c^{t},I) \leftarrow ED(z_t^t, c^t, I, I', w_{t}^{cfg}, w_{t}^{fg})$ \hfill Eq.~\ref{eq:cfg_edit_fg}
            \STATE $z_{t-1}^s \leftarrow \text{Sample}(z_{t}^s, \tilde{\epsilon}_\theta(z_t^s, c^{s}))$ \hfill Eq.~\ref{eq:ddim_reverse_recon}
            \STATE $z_{t-1}^t \leftarrow \text{Sample}(z_{t}^t, \tilde{\epsilon}_\theta(z_t^t, c^{t}))$ \hfill Eq.~\ref{eq:ddim_reverse_edit}
        \ENDFOR
        \end{algorithmic}
        \label{alg:fgs}
        \textbf{Return} $z_0^s, z_0^t$
\end{algorithm}
\begin{figure*}\centering
    \includegraphics[width=\linewidth]{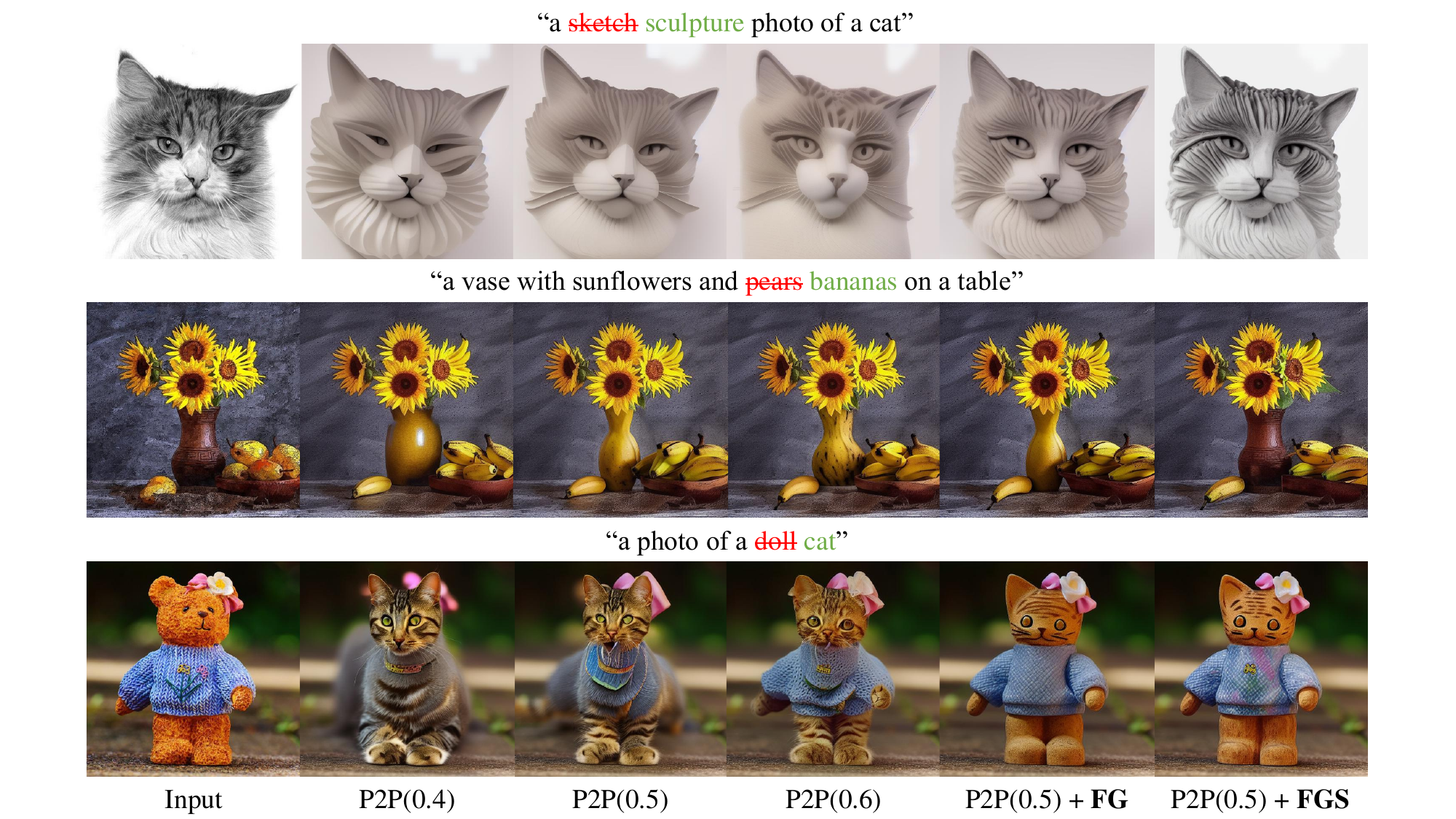}
    \vspace{-15pt}
    \caption{Qualitative comparison. Examples of image edits using P2P with different values of $\tau$ and with our proposed FG and FGS methods. In each row, the progression from left to right shows how varying $\tau$ or applying FG and FGS affects the trade-off between editability and faithfulness. The results indicate that FGS enhances faithfulness while maintaining editability, achieving more precise edits across different scenarios.}
    \vspace{-10pt}
    \label{fig:4_qual_comp}
\end{figure*}

\section{Experiments}
\label{sec:experiments}

\subsection{Quantitative comparison}
To demonstrate the effectiveness of our FGS, we use PIE-Bench~\cite{ju2024direct} for evaluation. Our primary comparisons are based on P2P~\cite{hertzprompt}. Directly applying the DDIM inversion~\cite{songdenoising} to reconstruct the input image makes it challenging to achieve precise reconstruction~\cite{mokady2023null}. To address this, we incorporate NMG~\cite{chonoise} to improve input image reconstruction. Since P2P transfers the layout of the input image, we use the structure distance (SD)~\cite{tumanyan2022splicing} between the edited image and the input image to measure faithfulness. For a more precise assessment, we calculate SD for both the entire image and the unedited region, identified using the mask provided by PIE-Bench. To evaluate editability, we use the CLIP score~\cite{hessel2021clipscore} between the target text and the edited image. Similarly, we calculate the CLIP score for both the entire image and the edited region, as specified by the PIE-Bench mask.

\begin{table}
    \centering
    \begin{tabular}{c|cc|cc}
    \hline
    \multirow{2}{*}{\textbf{Edit}} & \multicolumn{2}{c|}{\textbf{SD$\downarrow$}} & \multicolumn{2}{c}{\textbf{CLIP$\uparrow$}} \\
    & Whole & Unedited & Whole & Edited \\
    \hline
    P2P(0.4) & 33.70 & 13.38 & \textbf{26.11} & \textbf{23.19}  \\
    P2P(0.5) & 26.26 & 11.00 & 25.83 & 22.76 \\
    P2P(0.6) & 21.25 & 8.91 & 25.33 & 22.28 \\
    \hline
    \textbf{P2P(0.5)+FG} & 25.80 & 10.57 & 25.53 & 22.65 \\
    \textbf{P2P(0.5)+FGS} & \textbf{19.55} & \textbf{8.35} & 25.34 & 22.41 \\
    \hline
    \end{tabular}
    \vspace{-5pt}
    \caption{Quantitative comparison of image editing results. The number in brackets indicates the P2P timestep parameter $\tau$.}
    \vspace{-15pt}
    \label{table:1_quant_comp}
\end{table}

In Tab.\ref{table:1_quant_comp}, we present a quantitative comparison between the baseline and our method. In P2P, the injection of $I$ is controlled by the timestep parameter $\tau$. When $\tau$ is set to 0.5, $I$ is injected from $T$ to $0.5T$ during editing, meaning that a larger $\tau$ results in more $I$ being transferred from the reconstruction path to the editing path. As shown in Tab.\ref{table:1_quant_comp}, increasing $\tau$ decreases the SD, indicating higher faithfulness, while also lowering the CLIP score, reflecting reduced editability. This trend highlights the trade-off between editability and faithfulness. A comparison between the second and fourth rows of Tab.\ref{table:1_quant_comp} shows that introducing our FG improves faithfulness (i.e., reduces SD). However, due to misalignment between the CFG and FG directions, this improvement is limited. By additionally applying a scheduling method to implement our full approach (FGS), we achieve a more substantial increase in faithfulness. Notably, without our method, improving faithfulness would require increasing $\tau$ from 0.5 to 0.6, which would compromise editability. In contrast, our method achieves greater faithfulness with a minor reduction in editability than simply increasing $\tau$ from 0.5 to 0.6. Additionally, we integrate FGS with an alternative baseline and conduct a quantitative comparison; see the supplementary material for further details.

\begin{figure*}\centering
    \includegraphics[width=\linewidth]{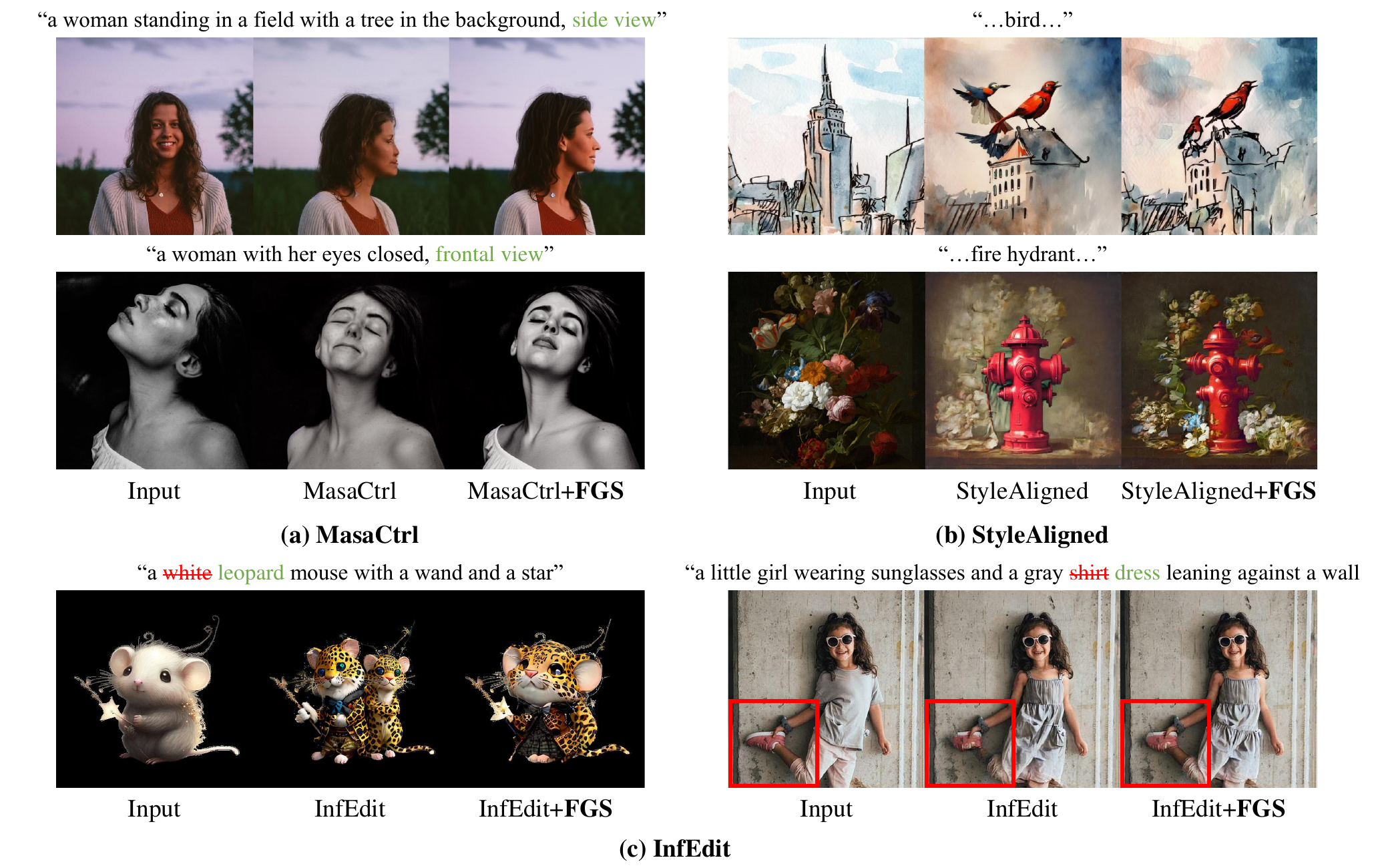}
    \vspace{-20pt}
    \caption{Qualitative comparison of various editing techniques with and without our proposed FGS. (a) MasaCtrl~\cite{cao2023masactrl}, (b) StyleAligned~\cite{hertz2024style}, and (c) InfEdit~\cite{xu2024inversion}. Results show that FGS improves the preservation of transferred information across all methods.}
    \vspace{-10pt}
    \label{fig:5_add_edit}
\end{figure*}

\subsection{Qualitative comparison}
Fig.\ref{fig:4_qual_comp} shows a qualitative comparison between P2P and P2P with our FGS. By controlling $\tau$, we can adjust the trade-off between editability and faithfulness. In the first row of Fig.~\ref{fig:4_qual_comp}, decreasing $\tau$ makes the image more sculptural but less similar to the original input. Conversely, increasing $\tau$ restores the original cat's features while diminishing the sculptural characteristics. This demonstrates how $\tau$ influences the trade-off between editability and faithfulness, making it challenging to achieve an optimal balance where both the original cat's features and the sculptural attributes are well-preserved. However, with our FGS, this balance is more effectively maintained. When editability is sufficient (e.g., $\tau=0.5$), FGS enhances faithfulness without significantly compromising editability, leading to more precise and controlled edits.

In some cases, adjusting $\tau$ alone is insufficient for achieving the desired level of faithfulness. For example, in the second row of Fig.~\ref{fig:4_qual_comp}, directly applying P2P changes the color and shape of the vase. Increasing $\tau$ to improve faithfulness does not fully restore the vase's original color and shape. With FGS, however, we successfully preserve the vase’s color and shape while changing the pears to bananas. IIn the third row, maintaining the structure of the doll's body is challenging without FGS, but with FGS, the overall structure is better preserved compared to the baseline. Some examples show that faithfulness improves with FG alone. However, when combined with the scheduling method, misalignment is further reduced, resulting in more consistent improvements in faithfulness.

\begin{figure*}\centering
    \includegraphics[width=\linewidth]{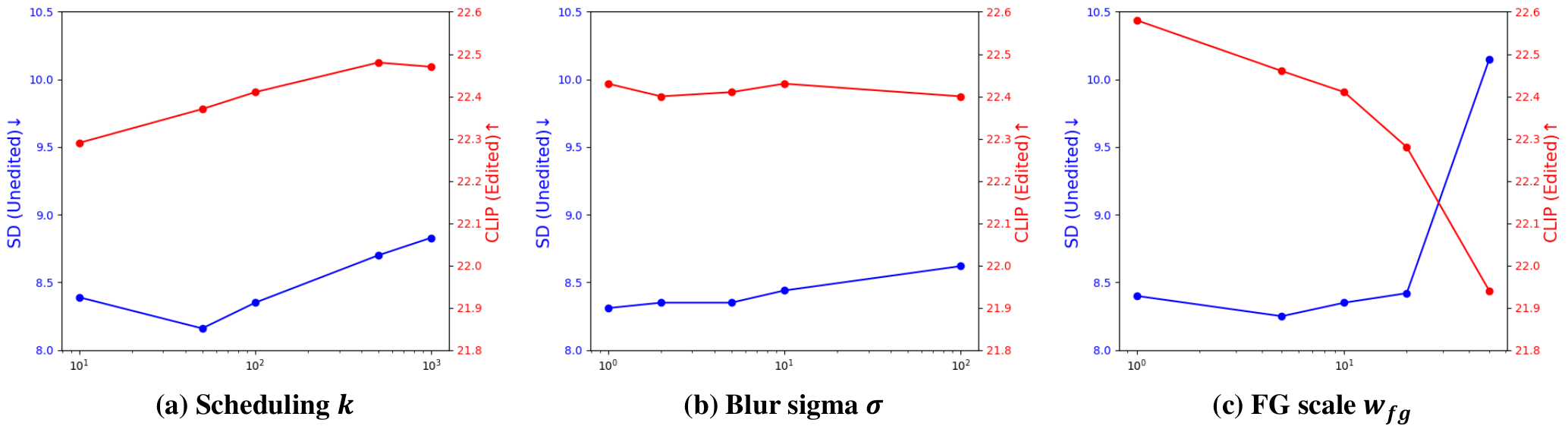}
    \vspace{-15pt}
    \caption{Ablation study of hyperparameter. (a) The impact of the scheduling parameter $k$ is evaluated using values $k \in \{10,50,100,500,1000\}$. (b) The influence of the Gaussian blur parameter $\sigma$ is tested across values $\sigma \in \{1,2,5,10,100\}$. (c) The effect of varying the FG scale $w_{fg}$ is examined with values $w_{fg} \in \{0,5,10,20,50\}$.}
    \label{fig:6_abl_control}
    \vspace{-5pt}
\end{figure*}

\subsection{Additional editing methods}
To demonstrate the versatility of FGS, we integrate our method with various editing techniques. For non-rigid editing, we apply MasaCtrl~\cite{cao2023masactrl}; for style transfer, we use StyleAligned~\cite{hertz2024style}; and for applicability to few-step distilled diffusion models such as LCM~\cite{luo2023latent}, we employ InfEdit~\cite{xu2024inversion}. In MasaCtrl, the overall shape of the image is generated based on the text prompt, while the details of the input image is transferred through the reconstruction path. Therefore, we set $w_{fg}$ to $w_i$. For StyleAligned, which focuses on transferring style, we set $w_{fg}$ to $w_i$. In Infedit, we perturb layout-related features to enhance the structural integrity of the image, so we set $w_{fg}$ to $w_d$.

Fig.~\ref{fig:5_add_edit} presents qualitative results for these editing methods with and without FGS. In MasaCtrl, without FGS, the input image content is not accurately preserved; however, with FGS, the identity of the woman is much better maintained. In StyleAligned, without FGS, the style components are transferred imprecisely. Applying FGS allows for more effective retention of the input image's style elements. In InfEdit, FGS enables better preservation of structural features than the baseline output. These results demonstrate that FGS can selectively enhance the transferred information from the input image, not only preserving structural features but also finer details such as style and identity. Furthermore, our method seamlessly integrates with few-step distilled diffusion models, such as LCM.

\begin{table}
    \centering
    \begin{tabular}{l|cc|cc}
    \hline
    \multirow{2}{*}{\textbf{Perturbation}} & \multicolumn{2}{c|}{\textbf{SD$\downarrow$}} & \multicolumn{2}{c}{\textbf{CLIP$\uparrow$}} \\
    & Whole & Unedited & Whole & Edited \\
    \hline
    - & 26.26 & 11.00 & \textbf{25.83} & \textbf{22.76} \\
    \hline
    Noise & 19.94 & 8.41 & 25.45 & 22.58 \\
    Identity & 20.27 & 8.69 & 25.50 & 22.46 \\
    Blur & \textbf{19.55} & \textbf{8.35} & 25.34 & 22.41 \\
    \hline
    \end{tabular}
    \vspace{-5pt}
    \caption{Ablation study of perturbation method.}
    \vspace{-15pt}
    \label{table:2_abl_perturb}
\end{table}

\subsection{Ablation study}
\paragraph{Perturbation method}
To approximate the unconditional output when treating the transferred information as the condition for the editing path, we applied perturbations to the transferred information. Tab.\ref{table:2_abl_perturb} presents experimental results using different perturbation methods with P2P as the editing technique. In addition to Gaussian blur, we tested adding Gaussian noise and replacing the attention matrix with an identity matrix, as proposed in PAG\cite{ahn2024self}. All perturbation methods improve faithfulness, as indicated by the reduction in SD. These results demonstrate that perturbing transferred information effectively approximates the unconditional output in CFG-like guidance. However, since our primary goal is to enhance faithfulness, we select Gaussian blur as the perturbation method.
\vspace{-10pt}
\paragraph{Hyperparameter selection}
As FGS is introduced, several several hyperparameters are required to control its behavior. Specifically, the scheduling parameter $k$ is introduced to regulate the rate of change in the guidance scales of CFG and FG. Additionally, Gaussian blur is adopted as the perturbation method, necessitating the selection of an appropriate blur amount. With the introduction of FG, the FG scale also be determined to balance faithfulness and editability effectively. Fig.~\ref{fig:6_abl_control} presents the performance variations across different hyperparameter settings. P2P is used as the editing method, and the results are reported in terms of SD for the unedited region and the CLIP score for the edited region. To systematically analyze the impact of these hyperparameters, we conducted ablation studies by varying each parameter while keeping the others fixed at their default values: $k=100$, $\sigma=5$, and $w_{fg}=10$.

As shown in Fig.~\ref{fig:6_abl_control}, variations in $k$ and $\sigma$ resulted in relatively minor performance changes, indicating that FGS is robust to moderate adjustments in these parameters. However, changes in the FG scale $w_{fg}$ had a more pronounced effect, significantly influencing the performance. These results suggest that while scheduling and blur perturbation provide stability, the FG scale plays a critical role in controlling the quality of the edited results. 
\vspace{-5pt}
\section{Conclusion}\label{sec:conclusion}
\vspace{-5pt}
In this work, we tackle the challenge of achieving optimal results in diffusion-based image editing. Editability defines the extent to which an image can be modified to match a target prompt, while faithfulness ensures the preservation of unaltered elements of the input image. The inherent trade-off between these two aspects poses a significant challenge, as improving one often comes at the expense of the other, making it difficult to achieve high-quality edits. To address this issue, we propose FGS, a novel framework designed to enhance faithfulness while minimizing its impact on editability. FGS combines a faithfulness guidance mechanism to better preserve input image information with a scheduling strategy to resolve the misalignment of editability and faithfulness. Our experimental results demonstrate that FGS improves faithfulness without significantly sacrificing editability, enabling more precise image edits. Furthermore, FGS is compatible with various editing techniques, enhancing editing quality across diverse tasks without introducing significant computational overhead. This work highlights the potential of FGS to advance real image editing by effectively managing the critical trade-offs that constrain diffusion-based image editing methods.

{
    \small
    \bibliographystyle{ieeenat_fullname}
    \bibliography{main}
}

\clearpage
\setcounter{table}{3}
\setcounter{figure}{6}
\setcounter{equation}{12}
\maketitlesupplementary

\appendix

\section{Implementation detail} 
In Prompt-to-Prompt (P2P)\cite{hertzprompt}, self-attention map in the middle block is perturbed to preserve the overall layout of the input image. MasaCtrl\cite{cao2023masactrl} perturbs the key and value components of self-attention transferred from the reconstruction path, while StyleAligned~\cite{hertz2024style} perturbs concatenated features transferred from the reconstruction of the input reference style. InfEdit~\cite{xu2024inversion}, similar to P2P, perturbs the query components of self-attention in the middle block to preserve the image layout. To prevent the collapse of the reconstruction path, both P2P and MasaCtrl incorporate NMG\cite{chonoise}. We use the Stable Diffusion (SD) 1.4~\cite{rombach2022high} for P2P and MasaCtrl, the SDXL~\cite{podellsdxl} for StyleAligned, and the LCM~\cite{luo2023latent} for InfEdit.
l experiments are conducted with a scheduling parameter of $k=100$ and a Gaussian blur intensity of $\sigma=5$. The FG scale $w_{fg}$ is set to 10 for P2P and MasaCtrl, 2.3 for InfEdit, and 3 for StyleAligned.

\section{Additional experiment}
\subsection{Quantitative comparison}
FGS improves faithfulness in image editing while minimizing editability loss. Accordingly, our primary evaluation focuses on comparing baseline models with and without FGS integration. Table~\ref{table:4_add_quant_comp} presents an additional quantitative evaluation of models with and without FGS. We include P2P with SPDInv~\cite{li2024source}, a state-of-the-art inversion method, as well as InfEdit, which uses an inversion method based on virtual inversion (VI)\cite{xu2024inversion} and an editing method with unified attention control (UAC)\cite{xu2024inversion}. Furthermore, we extend our comparison with instruction-based methods, a widely adopted approach for image editing with diffusion models. Unlike reconstruction-based models, instruction-based methods require additional training to edit real images. As shown in Table~\ref{table:4_add_quant_comp}, applying FGS to SPDInv and InfEdit, similar to P2P with NMG, enhances faithfulness with minimal impact on editability. Moreover, our results indicate that instruction-based editing methods generally perform less than reconstruction-based approaches, despite requiring additional training.

\subsection{Qualitative comparison}
\begin{figure*}[h]\centering
    \includegraphics[width=\linewidth]{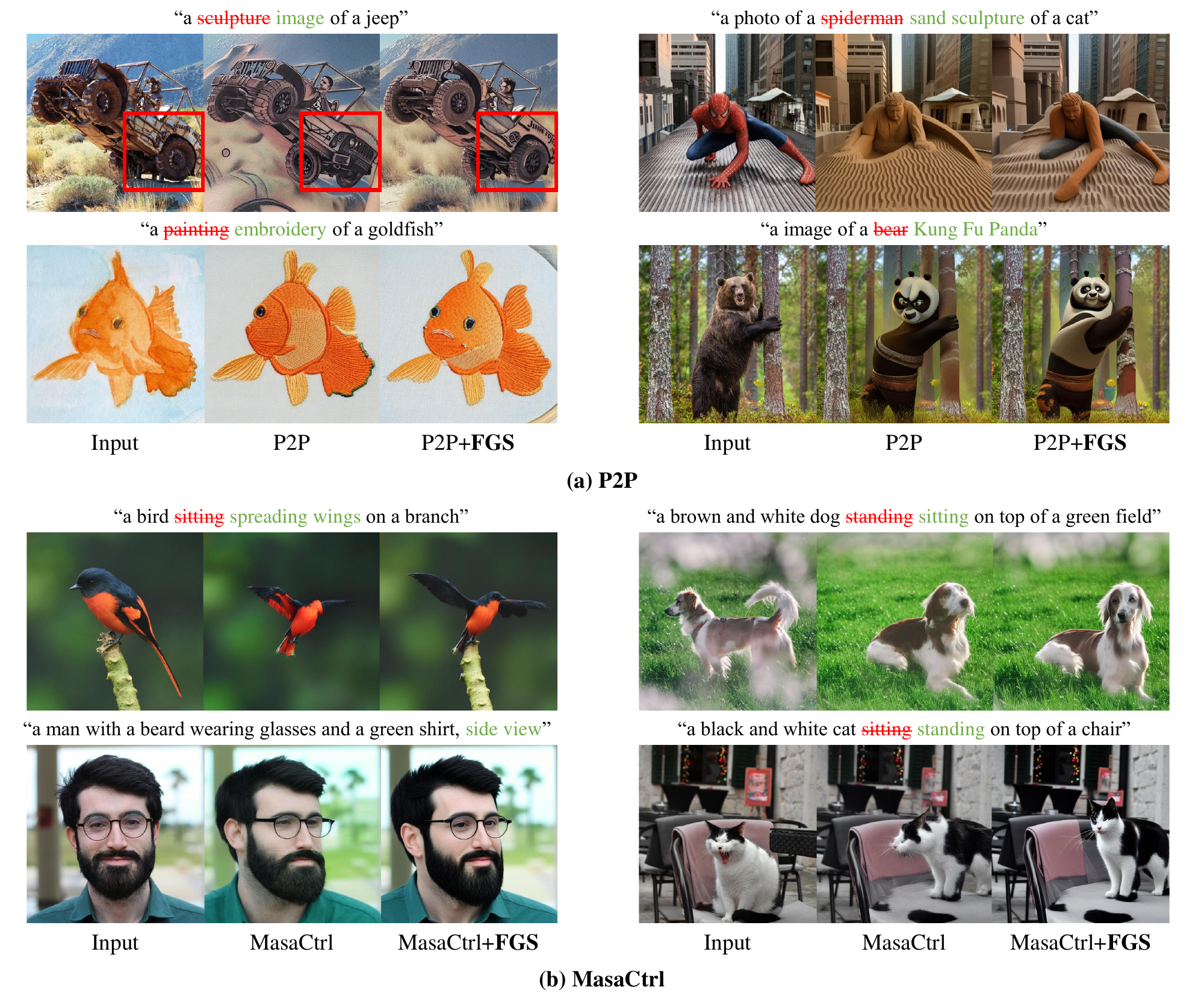}
    \vspace{-10pt}
    \caption{Additional qualitative comparison on P2P and MasaCtrl}
    \label{fig:8_add_qual_compt}
\end{figure*}

\begin{figure*}[h]\centering
    \includegraphics[width=\linewidth]{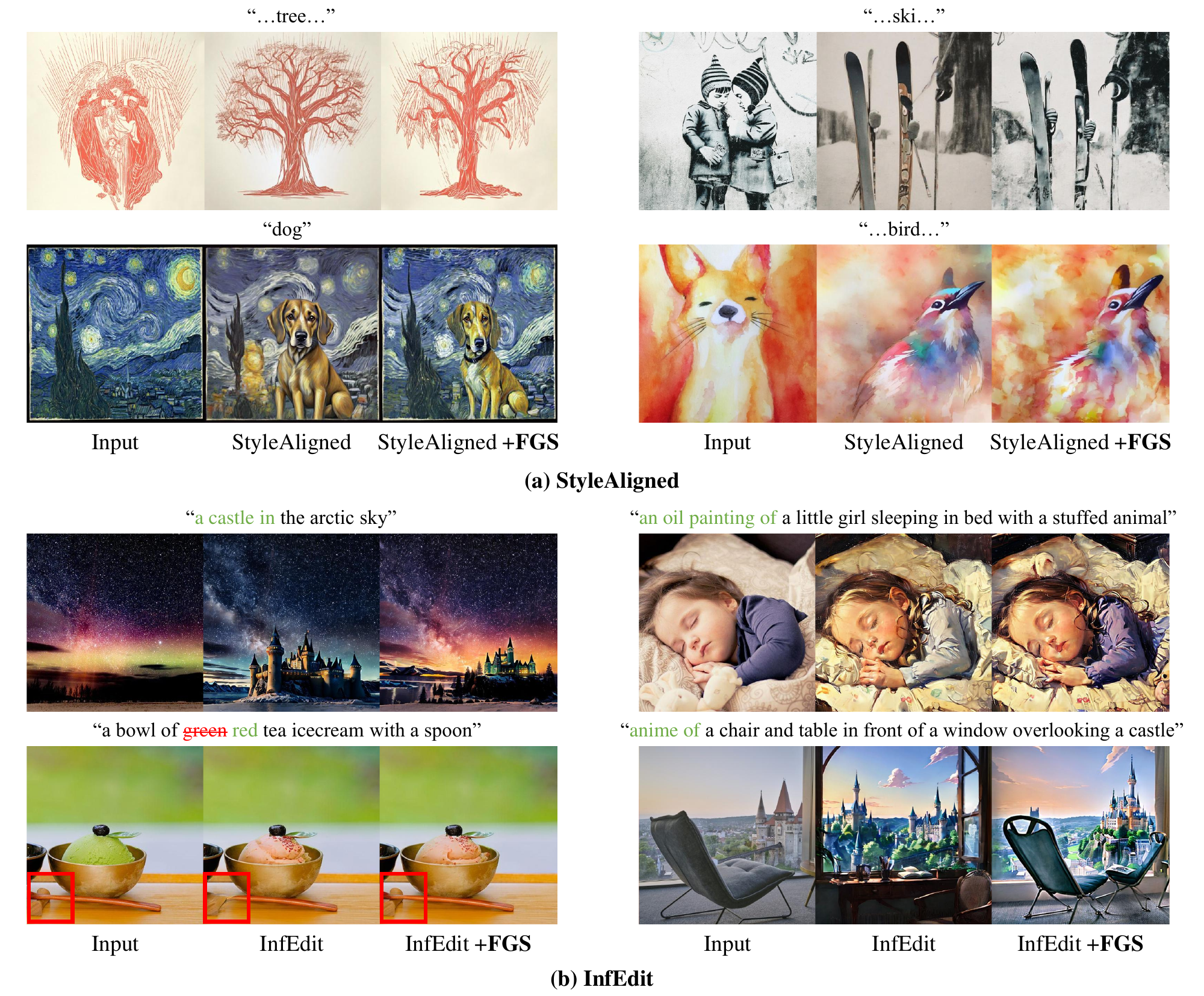}
    \vspace{-10pt}
    \caption{Additional qualitative comparison on StyleAligned and InfEdit}
    \label{fig:9_add_qual_compt}
\end{figure*}

We present additional qualitative comparison results of Fig.~\ref{fig:8_add_qual_compt} and Fig.~\ref{fig:9_add_qual_compt}. As illustrated in the figure, our FGS enables the edited images to preserve the overall input image content and style better than the baseline methods.

\section{Limitations}
\begin{figure*}[h]\centering
    \includegraphics[width=\linewidth]{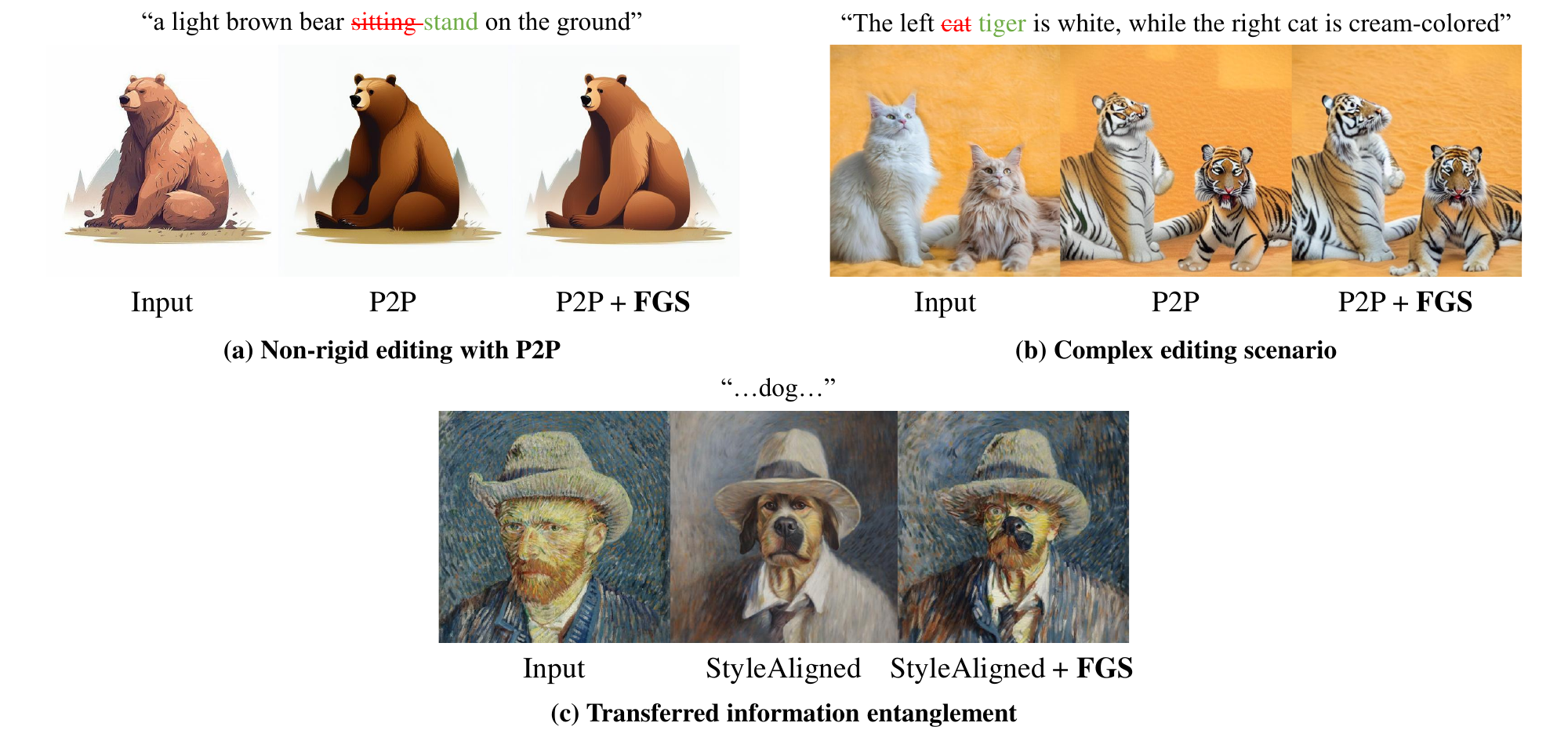}
    \vspace{-10pt}
    \caption{Limitation examples}
    \label{fig:7_limitation}
\end{figure*}

FGS builds upon a baseline editing method, inheriting its limitations. As shown in Fig.~\ref{fig:7_limitation}(a), P2P, which is designed for rigid editing, struggles with non-rigid editing such as pose changes, even when combined with FGS. Similarly, as seen in Fig.~\ref{fig:7_limitation}(b), editing complex scenarios remains challenging. Accurately distinguishing between the left and right cats is crucial in the example. However, since diffusion models have difficulty understanding complex images, precise editing becomes challenging. Lastly, enhancing the transferred information can degrade the editing results if the transferred information is not well disentangled. As shown in Fig.~\ref{fig:7_limitation}(c), when the transferred information is poorly disentangled, StyleAligned generates an output where the dog retains the shape of the input image. As FGS amplifies the transferred information, it fails to generate a realistic dog image.

Additionally, FGS increases memory usage by approximately 3GB and inference time by about 4 seconds per image. However, it remains manageable on a single RTX 4090 GPU. More importantly, it enhances faithfulness without additional training.

\begin{table*}
    \centering
    \begin{tabular}{c|c|c|c|cc|cc}
    \hline
    & \multirow{2}{*}{\textbf{Inv}} & \multirow{2}{*}{\textbf{Edit}} & \multirow{2}{*}{\textbf{FGS}} & \multicolumn{2}{c|}{\textbf{SD$\downarrow$}} & \multicolumn{2}{c}{\textbf{CLIP$\uparrow$}} \\
    & & & & Whole & Unedited & Whole & Edited \\
    \hline
    \multirow{6}{*}{Reconstruction-based} & NMG~\cite{li2024source} & P2P~\cite{hertzprompt} & X & 26.26 & 11.00 & \textbf{25.83} & \textbf{22.76} \\
    & NMG~\cite{li2024source} & P2P~\cite{hertzprompt} & O & \textbf{19.55} & \textbf{8.35} & 25.34 & 22.41 \\
    \cline{2-8}
    & SPDInv~\cite{li2024source} & P2P~\cite{hertzprompt} & X & 10.60 & 4.00 & \textbf{24.73} & \textbf{21.72} \\
    & SPDInv~\cite{li2024source} & P2P~\cite{hertzprompt} & O & \textbf{8.53} & \textbf{3.71} & 24.10 & 21.30 \\
    \cline{2-8}
    & VI~\cite{xu2024inversion} & UAC~\cite{xu2024inversion} & X & 29.98 & 8.86 & \textbf{26.04} & \textbf{23.26} \\
    & VI~\cite{xu2024inversion} & UAC~\cite{xu2024inversion} & O & \textbf{26.67} & \textbf{8.35} & 25.86 & 23.11 \\
    \noalign{\hrule height 3pt}
    \multirow{3}{*}{Instruction-based} & \multicolumn{3}{c|}{InstructPix2Pix~\cite{brooks2023instructpix2pix}} & \textbf{56.52} & 33.81 & 23.60 & \textbf{21.77} \\
    \cline{2-8}
    & \multicolumn{3}{c|}{MagicBrush~\cite{zhang2023magicbrush}} & 69.10 & \textbf{19.67} & \textbf{23.63} & 21.27 \\
    \cline{2-8}
    & \multicolumn{3}{c|}{InstDiff~\cite{geng2024instructdiffusion}} & 79.11 & 37.95 & 23.07 & 21.15 \\
    \hline
    \end{tabular}
    \caption{Additional quantitative comparison on PIE-Bench. For reconstruction-based editing methods, bold font indicates the best performance both with and without FGS. For instruction-based methods, bold font highlights the best performance among instruction-based approaches.}
    \label{table:4_add_quant_comp}
\end{table*}

\end{document}